\documentclass[twoside]{article}

\usepackage[accepted]{aistats2023}
\usepackage{graphicx}
\usepackage{amssymb}
\usepackage{booktabs}
\usepackage{natbib}
\usepackage{hyperref}
\usepackage{wrapfig}
\usepackage{enumitem}

\usepackage{algorithm}

\usepackage{amsmath}
\usepackage[noend]{algpseudocode}
\usepackage[switch]{lineno}
\usepackage{subcaption}

\begin{document}



\twocolumn[

\aistatstitle{Positional Encoder Graph Neural Networks for Geographic Data}

\aistatsauthor{ Konstantin Klemmer \And Nathan Safir \And  Daniel B. Neill }

\aistatsaddress{ Microsoft Research \And  University of Georgia \And New York University } ]

\begin{abstract}
    Graph neural networks (GNNs) provide a powerful and scalable solution for modeling continuous spatial data. However, they often rely on Euclidean distances to construct the input graphs. This assumption can be improbable in many real-world settings, where the spatial structure is more complex and explicitly non-Euclidean (e.g., road networks). Here, we propose PE-GNN, a new framework that incorporates spatial context and correlation explicitly into the models. Building on recent advances in geospatial auxiliary task learning and semantic spatial embeddings, our proposed method (1) learns a context-aware vector encoding of the geographic coordinates and (2) predicts spatial autocorrelation in the data in parallel with the main task. On spatial interpolation and regression tasks, we show the effectiveness of our approach, improving performance over different state-of-the-art GNN approaches. We observe that our approach not only vastly improves over the GNN baselines, but can match Gaussian processes, the most commonly utilized method for spatial interpolation problems.
\end{abstract}

\section{Introduction}
Geographic data is characterized by a natural geometric structure, which often defines the observed spatial pattern. While traditional neural network approaches do not have an intuition to account for spatial dynamics, graph neural networks (GNNs) can represent spatial structures graphically. The recent years have seen many applications leveraging GNNs for modeling tasks in the geographic domain, such as inferring properties of a point-of-interest \citep{Zhu2020} or predicting the speed of traffic at a certain location \citep{Chen2019}. Nonetheless, as we show in this study, GNNs are not necessarily sufficient for modeling complex spatial effects: spatial context can be different at each location, which may be reflected in the relationship with its spatial neighborhood. The study of spatial context and dependencies has attracted increasing attention in the machine learning community, with studies on spatial context embeddings \citep{Mai2020a,Yin2019} and spatially explicit auxiliary task learning \citep{Klemmer2021a}.

Here, we seek to merge these streams of research. We propose the positional encoder graph neural network (\textbf{PE-GNN}), a flexible approach for better encoding spatial context into GNN-based predictive models. \textbf{PE-GNN} is highly modular and can work with any GNN backbone. It contains a positional encoder (PE) \citep{Vaswani2017,Mai2020a}, which learns a contextual embedding for point coordinates throughout training. The embedding returned by PE is concatenated with other node features to provide the training data for the GNN operator. \textbf{PE-GNN} further predicts the local spatial autocorrelation of the output as an auxiliary task in parallel to the main objective, expanding the approach proposed by \cite{Klemmer2021a} to continuous spatial coordinates. We train \textbf{PE-GNN} by constructing a novel training graph, based on $k$-nearest-neighborhood, from a randomly sampled batch of points at each training step. This forces PE to learn generalizable features, as the same point coordinate might have different spatial neighbors at different training steps. Distances between nodes are reflected as edge weights. This training approach also leads us to compute a ``shuffled" Moran's I, implicitly nudging the model to learn a general representation of spatial autocorrelation which works across varying neighbor sets. Over a range of spatial regression tasks, we show that \textbf{PE-GNN} consistently improves performance of different GNN backbones.

Our contributions can be summarized as follows:
\begin{itemize}%[leftmargin=*]
    \item We propose \textbf{PE-GNN}, a novel GNN architecture including a positional encoder learning spatial context embeddings for each point coordinate to improve predictions.
    \item We propose a novel way of training the positional encoder (PE): While \cite{Mai2020a} train PE in an unsupervised fashion and \cite{Mai2020b} use PE in a joint embedding with a data-dependent, secondary encoder (e.g., text encoder), we use the output of PE concatenated with other node features to directly predict an outcome variable. PE learns through backpropagation on the main regression loss in an end-to-end fashion. Training PE thus takes into account not only the eventual variable of interest, but also further contextual information at the current location--and its relation to other points. Within \textbf{PE-GNN}, spatial information is thus represented both through the constructed graph and the learned PE embeddings. 
    \item We expand the Moran's I auxiliary task learning framework proposed by \cite{Klemmer2021a} for continuous spatial coordinates.
    \item Our training strategy involves the creation of a new training graph at each training step from the current, random point batch. This enables learning of a more generalizable PE embedding and allows computation of a ``shuffled" Moran's I, which accounts for different neighbors at different training steps, thus tackling the well-known scale sensitivity of Moran's I.
    \item To the best of our knowledge, \textbf{PE-GNN} is the first GNN based approach that is competitive with Gaussian Processes on pure spatial interpolation tasks, i.e., predicting a (continuous) output based solely on spatial coordinates, as well as substantially improving GNN performance on all predictive tasks.
\end{itemize}

\section{Related work}
\subsection{Traditional and neural-network-based spatial regression modeling}

Our work considers the problem of modeling geospatial data. This poses a distinct challenge, as standard regression models (such as OLS) fail to address the spatial nature of the data, which can result in spatially correlated residuals. To address this, spatial lag models \citep{anselin2001spatial} add a spatial lag term to the regression equation that is proportional to the dependent variable values of nearby observations, assigned by a weight matrix. Likewise, kernel regression takes a weighted average of nearby points when predicting the dependent variable. The most popular off-the-shelf methods for modeling continuous spatial data are based on Gaussian processes \citep{Datta2016}. Recently, there has been a rise of research on applications of neural network models for spatial modeling tasks. More specifically, graph neural networks (GNNs) are often used for these tasks with the spatial data represented graphically. Particularly, they offer flexibility and scalability advantages over traditional spatial modeling approaches. Specific GNN operators including Graph Convolutions \citep{Kipf2017}, Graph Attention \citep{Velickovic2018} and GraphSAGE \citep{Hamilton2017a} are powerful methods for inference and representation learning with spatial data. Recently, GNN approaches tailored to the specific complexities of geospatial data have been developed. The authors of Kriging Convolutional Networks \citep{Appleby2020} propose using GNNs to perform a modified kriging task. \cite{Hamilton2017a} apply GNNs for a spatio-temporal Kriging task, recovering data from unsampled nodes on an input graph. We look to extend this line of research by providing stronger, explicit capacities for GNNs to learn spatial structures. Additionally, our proposed method is highly modular and can be combined with any GNN backbone.

\subsection{Spatial context embeddings for geographic data}

Through many decades of research on spatial patterns, a myriad of measures, metrics, and statistics have been developed to cover a broad range of spatial interactions. All of these measures seek to transform spatial locations, with optional associated features, into some meaningful embedding, for example, a theoretical distribution of the locations or a measure of spatial association. 
%A set of point locations in $2d$-space may be distributed randomly, or may follow some underlying spatial pattern. Spatial point processes are some of the oldest methods of analyzing spatial data and are a useful tool when point locations depend on one another, e.g., in the case of epidemiological or ecological data. Point processes are particularly popular for modeling geographic data, for example Cox processes \cite{Aglietti2019} and Hawkes processes \cite{Unwin2021}. Empirical observations can be embedded using these point processes, for example, to make predictions on future spatial spread. If we now assume that point locations also come with a continuous feature value, the range of potential spatial effects and associated metrics is vastly expanded. 
The most common metric for continuous geographic data is the Moran's I statistic, developed by \cite{Anselin1995}. Moran's I measures local and global spatial autocorrelation and acts as a detector of spatial clusters and outliers. The metric has also motivated several methodological expansions, like local spatial heteroskedasticity \citep{Ord2012} and local spatial dispersion \citep{Westerholt2018}. Measures of spatial autocorrelation have already been shown to be useful for improving neural network models through auxiliary task learning \citep{Klemmer2021a}, model selection \citep{Klemmer2019a}, embedding losses \citep{Klemmer2021d} and localized representation learning \citep{Fu2019}. Beyond these traditional metrics, recent years have seen the emergence of neural network based embeddings for geographic information. \cite{Wang2017a} use kernel embeddings to learn social media user locations. \cite{Fu2019} devise an approach using local point-of-interest (POI) information to learn region embeddings and integrate similarities between neighboring regions to learn mobile check-ins. \cite{Yin2019} develop GPS2Vec, an embedding approach for latitude-longitude coordinates, based on a grid cell encoding and spatial context (e.g., tweets and images). \cite{Mai2020a} developed Space2Vec, another latitude-longitude embedding without requiring further context like tweets or POIs. Space2Vec transforms the input coordinates using  sinusoidal functions and then reprojects them into a desired output space using linear layers. In follow-up work, \cite{Mai2020b} first propose the direct integration of Space2Vec into downstream tasks and show its potential with experiments on spatial semantic lifting and geographic question answering. In this study, we propose to generalize their approach to any geospatial regression task by conveniently integrating Space2Vec embeddings into GNNs.

\section{Method}
\begin{figure*}[ht]%{0.4\textwidth}
%\vskip -0.1in
\centering
\includegraphics[scale=0.5]{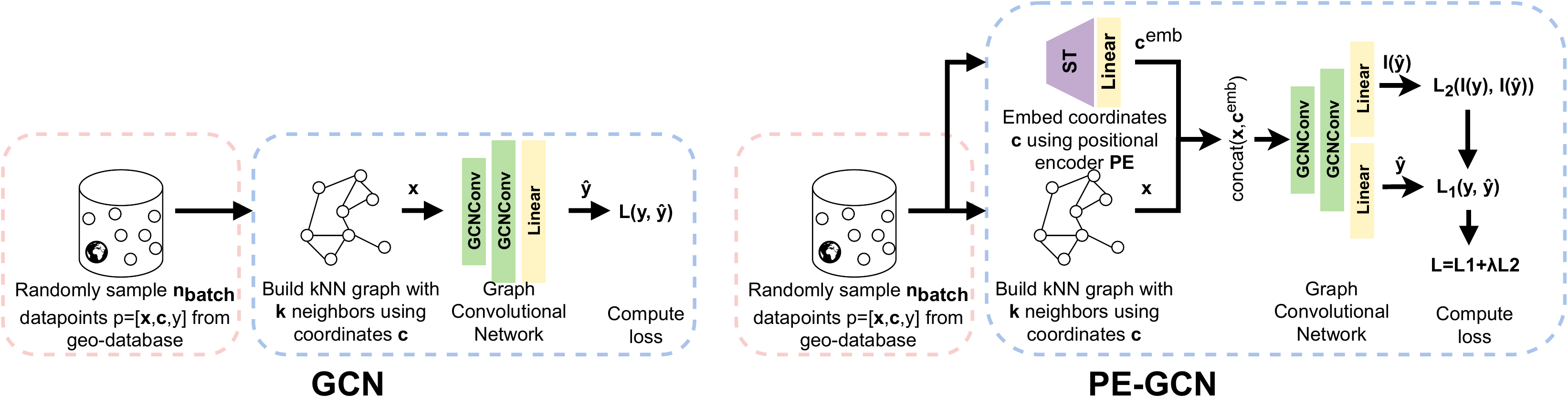}
%\includegraphics{fig1.pdf}
%\vskip -0.1in
\caption{\textbf{PE-GCN} compared to the \textbf{GCN} baseline: \textbf{PE-GCN} contains a (1) positional encoder network, learning a spatial context embedding throughout training which is concatenated with node-level features and (2) an auxiliary learner, predicting the spatial autocorrelation of the outcome variable simultaneously to the main regression task.}
\label{fig:1}
%\vskip -0.15in
%\vskip -0.5in
\end{figure*}

\subsection{Graph Neural Networks with Geographic Data}
 
We now present \textbf{PE-GNN}, using Graph Convolutional Networks (GCNs) as example backbone. Let us first define a datapoint $p_i = \{y_i,\mathbf{x_i},\mathbf{c_i}\}$, where $y_i$ is a continuous target variable (scalar), $\mathbf{x_i}$ is a vector of predictive features and $\mathbf{c_i}$ is a vector of point coordinates (latitude / longitude pairs). We use the great-circle distance $d_{ij} = haversin(\mathbf{c_i},\mathbf{c_j})$ between point coordinates to create a graph of all points in the set, using a $k$-nearest-neighbor approach to define each point's neighborhood. The graph $G = (V,E)$ consists of a set of vertices (or nodes) $V = \{v_1,\dots,v_n\}$ and a set of edges $E = \{e_1,\dots,e_m\}$ as assigned by the adjacency matrix $\mathbf{A}$. Each vertex $i \in V$ has respective node features $\mathbf{x_i}$ and target variable $y_i$. While the adjacency matrix $\mathbf{A}$ usually comes as a binary matrix (with values of $1$ indicating adjacency and values of $0$ otherwise), one can account for different distances between nodes and use point distances $d_{ij}$ or kernel transformations thereof \citep{Appleby2020} to weight $\mathbf{A}$. Given a degree matrix $\mathbf{D}$ and an identity matrix $\mathbf{I}$, the normalized adjacency matrix $\mathbf{\bar{A}}$ is defined as:
\begin{equation}
\label{eq:1}
    \mathbf{\bar{A}} = \mathbf{D}^{-1/2}(\mathbf{A} + \mathbf{I})\mathbf{D}^{-1/2}
\end{equation}

As proposed by \cite{Kipf2017}, a GCN layer can now be defined as:
\begin{equation}
\label{eq:2}
    \mathbf{H}^{(l)} = \sigma (\mathbf{\bar{A}} \mathbf{H}^{(l-1)}\mathbf{W}^{(l)}), l = 1,\dots,L
\end{equation}

\noindent where $\sigma$ describes an activation function (e.g., ReLU) and $\mathbf{W}^{(l)}$ is a weight matrix parametrizing GCN layer $l$. The input for the first GCN layer $\mathbf{H^{(0)}}$ is given by the feature matrix $\mathbf{X}$ containing all node feature vectors $\mathbf{x_1},\dots,\mathbf{x_n}$. The assembled GCN predicts the output $\mathbf{\hat{Y}} = GCN(\mathbf{X},\Theta_{GCN})$ parametrized by $\Theta_{GCN}$.

\subsection{Context-aware spatial coordinate embeddings}

Traditionally, the only intuition for spatial context in GCNs stems from connections between nodes which allow for graph convolutions, akin to pixel convolutions with image data. This can restrict the capacity of the GCN to capture spatial patterns: While defining good neighborhood structures can be crucial for GCN performance, this often comes down to somewhat arbitrary choices like selecting the $k$ nearest neighbors of each node. Without prior knowledge on the underlying data, the process of setting the right neighborhood parameters may require extensive testing. Furthermore, a single value of $k$ might not be best for all nodes: different locations might be more or less dependent on their neighbors. Assuming that no underlying graph connecting point locations is known, one would typically construct a graph using the distance (Euclidean or other) between pairs of points. In many real world settings (e.g., points-of-interest along a road network) this assumption is unrealistic and may lead to poorly defined neighborhoods. Lastly, GCNs contain no intrinsic tool to transform point coordinates into a different (latent) space that might be more informative for representing the spatial structure, with respect to the particular problem the GCN is trying to solve. 

As such, GCNs can struggle with tasks that explicitly require learning of complex spatial dependencies, as we confirm in our experiments. We propose a novel approach to overcome these difficulties, by devising a new positional encoder module, learning a flexible spatial context encoding for each geographic location. Given a batch of datapoints, we create the spatial coordinate matrix $\mathbf{C}$ from individual point coordinates $\mathbf{c_1},\dots,\mathbf{c_n}$ and define a positional encoder $PE(\mathbf{C},\sigma_{min},\sigma_{max},\Theta_{PE}) = NN(ST(\mathbf{C},\sigma_{min}$, $\sigma_{max}),\Theta_{PE})$, consisting of a sinusoidal transform $ST(\sigma_{min}$, $\sigma_{max})$ and a fully-connected neural network $NN(\Theta_{PE})$, parametrized by $\Theta_{PE}$. Following the intuition of transformers \citep{Vaswani2017} for geographic coordinates \citep{Mai2020a}, the sinusoidal transform is a concatenation of scale-sensitive sinusoidal functions at different frequencies, so that

\begin{equation}
\label{eq:3}
\begin{aligned}
& ST(\mathbf{C},\sigma_{min},\sigma_{max}) = \\
& [ST_{0}(\mathbf{C},\sigma_{min},\sigma_{max});\dots;ST_{S-1}(\mathbf{C},\sigma_{min},\sigma_{max})]  
\end{aligned}
\end{equation}

\noindent with $S$ being the total number of grid scales and $\sigma_{min}$ and $\sigma_{max}$ setting the minimum and maximum grid scale (comparable to the lengthscale parameter of a kernel). The scale-specific encoder $ST_{s}(\mathbf{C},\sigma_{min},\sigma_{max}) = [ST_{s,1}(\mathbf{C},\sigma_{min},\sigma_{max});ST_{s,2}(\mathbf{C},\sigma_{min},\sigma_{max})]$ processes the spatial dimensions $v$ (e.g., latitude and longitude) of $\mathbf{C}$ separately, so that

\begin{equation}
\label{eq:4}
\begin{aligned}
& ST_{s,v}(\mathbf{C,\sigma_{min},\sigma_{max}}) = \\
& \left[\cos\left(\frac{\mathbf{C}^{[v]}}{\sigma_{min}g^{s / (S-1)}}\right);\sin\left(\frac{\mathbf{C}^{[v]}}{\sigma_{min}g^{s / (S-1)}}\right)\right]    \\
& \forall s\in\{0,\ldots, S-1\}, \forall v\in\{1,2\}, 
\end{aligned}
\end{equation}

\noindent where $g = \frac{\sigma_{max}}{\sigma_{min}}$.
The output from $ST$ is then fed through the fully connected neural network $NN(\Theta_{PE})$ to transform it into the desired vector space shape, creating the coordinate embedding matrix $\mathbf{C}_{emb} = PE(\mathbf{C},\sigma_{min},\sigma_{max},\Theta_{PE})$. 

\subsection{Auxiliary learning of spatial autocorrelation}

Geographic data often exhibit spatial autocorrelation: observations are related, in some shape or form, to their geographic neighbors. Spatial autocorrelation can be measured using the Moran's I metric of local spatial autocorrelation \citep{Anselin1995}. Moran's I captures localized homogeneity and outliers, functioning as a detector of spatial clustering and spatial change patterns. In the context of our problem, the Moran's I measure of spatial autocorrelation for outcome variable $y_{i}$ is defined as:

\begin{equation}
\label{eq:5}
\begin{split}
    I_{i} = (n-1) \frac{(y_{i} -\bar{y}_{i})}{\sum^{n}_{j=1} (y_{j} -\bar{y}_{j})^2} \sum^{n}_{j=1, j \neq i} a_{i,j} (y_{j} -\bar{y}_{j}),
\end{split}
\end{equation}

\noindent where $a_{i,j} \in \mathbf{A}$ denotes adjacency of observations $i$ and $j$. 

As proposed by \cite{Klemmer2021a}, predicting the Moran's I metric of the output can be used as auxiliary task during training. Auxiliary task learning \citep{Suddarth1990} is a special case of multi-task learning, where one learning algorithm tackles two or more tasks at once. In auxiliary task learning, we are only interested in the predictions of one task; however, adding additional, auxiliary tasks to the learner might improve performance on the primary problem: the auxiliary task can add context to the learning problem that can help solve the main problem. This approach is commonly used, for example in reinforcement learning \citep{Flet-Berliac2019} or computer vision \citep{Hou2019,Jaderberg2017}.

Translated to our GCN setting, we seek to predict the outcome $\mathbf{Y}$ and its local Moran's I metric $I(\mathbf{Y})$ using the same network, so that $[\hat{\mathbf{Y}},\hat{I(\mathbf{Y})}] = GCN(\mathbf{X})$. As \cite{Klemmer2021a} note, the local Moran's I metric is scale-sensitive and, due to its restriction to local neighborhoods, can miss out on longer-distance spatial effects \citep{Feng2019,Meng2014}. But while \cite{Klemmer2021a} propose to compute the Moran's I at different resolutions, the GCN setting allows for a different, novel approach to overcome this issue: Rather than constructing the graph of training points a priori, we opt for a procedure where in each training step, $n_{batch}$ points are sampled from the training data as batch $B$. A graph with corresponding adjacency matrix $\mathbf{A}_{B}$ is constructed for the batch and the Moran's I metric of the outcome variable $I(\mathbf{Y}_{B})$ is computed. This approach brings a unique advantage: When training with (randomly shuffled) batches, points may have different neighbors in different training iterations. The Moran's I for point $i$ can thus change throughout iterations, reflecting a differing set of more distant or closer neighbors. This also naturally helps to tackle Moran's I scale sensitivity. Altogether, we refer to this altered Moran's I as ``shuffled Moran's I". 

\subsection{Positional Encoder Graph Neural Network (PE-GNN)}

%\begin{figure*}[!ht]%{0.4\textwidth}
%\vskip -0.1in
%\centering
%\includegraphics[scale=0.53]{cali_housing_pred.png}
%\includegraphics{fig1.pdf}
%\vskip -0.1in
%\caption{California Housing dataset: Real values and predictions using GraphSAGE and PE-GraphSAGE}
%\label{fig:2}
%\vskip -0.15in
%\vskip -0.5in
%\end{figure*}

We now assemble the different modules of our method and introduce the Positional Encoder Graph Neural Network (\textbf{PE-GNN}). The whole modeling pipeline of \textbf{PE-GNN} compared to a naive GNN approach is pictured in Figure \ref{fig:1}. Sticking to the GCN example, \textbf{PE-GCN} is constructed as follows: Assuming a batch $B$ of randomly sampled points $p_{1},\dots,p_{n_{batch}} \in B$, a spatial graph is constructed from point coordinates $\mathbf{c}_{1}, \dots, \mathbf{c}_{n_{batch}}$ using $k$-nearest-neighborhood, resulting in adjacency matrix $\mathbf{A}_{B}$. The point coordinates are then subsequently fed through the positional encoder $PE(\Theta_{PE})$, consisting of the sinusoidal transform $ST$ and a single fully-connected layer with sigmoid activation, embedding the $2d$ coordinates in a custom latent space and returning vector embeddings $\mathbf{c}_{1}^{emb}, \dots, \mathbf{c}_{n_{batch}}^{emb} =\mathbf{C}_{B}^{emb}$. The neural network allows for explicit learning of spatial context, reflected in the vector embedding. We then concatenate the positional encoder output with the node features, to create the input for the first GCN layer:

\begin{equation}
\label{eq:6}
\mathbf{H}^{(0)} = concat(\mathbf{X}_{B},\mathbf{C}_{B}^{emb})
\end{equation}

The subsequent layers follow according to Equation \ref{eq:2}. Note here that this approach is distinctly different from \cite{Mai2020b}, who learn a specific joint embedding between the geographic coordinates and potential other inputs (e.g., text data). Our approach allows for separate treatment of geographic coordinates and potential other predictors, allowing a higher degree of flexibility: \textbf{PE-GCN} can be deployed for any regression task, geo-referenced in the form of latitude longitude coordinates. Lastly, to integrate the Moran's I auxiliary task, we compute the metric $I(\mathbf{Y}_{B})$ for our outcome variable $\mathbf{Y}_{B}$ at the beginning of each training step according to Equation \ref{eq:5}, using spatial weights from $\mathbf{A}_{B}$. Prediction is then facilitated by creating two prediction heads, here linear layers, while the graph operation layers (e.g., GCN layers) are shared between tasks. Finally, we obtain predicted values $\hat{\mathbf{Y}}_{B}$ and $\hat{I(\mathbf{Y}_{B})}$. The loss of \textbf{PE-GCN} can be computed with any regression criterion, for example mean squared error (MSE):

\begin{equation}
\label{eq:8}
\mathcal{L} = MSE(\hat{\mathbf{Y}}_{B}, \mathbf{Y}_{B}) + \lambda MSE(I(\hat{\mathbf{Y}}_{B}), I(\mathbf{Y}_{B}))
\end{equation}

where $\lambda$ denotes the auxiliary task weight. The final model is denoted as $M_{\Theta_{PE},\Theta_{GCN}}$. Algorithm \ref{alg:1} describes a training cycle.

\begin{algorithm}[h!]
	\caption{PE-GNN Training}\label{training}
	\label{alg:1}
	\begin{algorithmic}[1]
	    \Require{$M$, $\lambda$, $k$, $tsteps$,$n_{batch}$ hyper-parameter}
	    \State Initialize model $M$ with random weights and hyper-parameter
	    \State Set optimizer with hyper-parameter
		\For{number of training steps ($tsteps$)}
		\State Sample minibatch $B$ of $n_{batch}$ points with features $\mathbf{X}_{B}$, coordinates $\mathbf{C}_{B}$ and outcome $\mathbf{Y}_{B}$.
		\State Construct a spatial graph with adjacency matrix $\mathbf{A}_B$ from coordinates $\mathbf{C}_{B}$ using $k$-nearest neighbors
        \State Using spatial adjacency $\mathbf{A}_B$, compute Moran's I of output as $I(\mathbf{Y}_B)$
        \State Predict outcome 
        \Statex $[\hat{\mathbf{Y}}_{B},I(\hat{\mathbf{Y}}_{B})] = M_{\Theta_{PE},\Theta_{GCN}}(\mathbf{X}_{B},\mathbf{C}_{B},\mathbf{A}_{B})$
         \State Compute loss \Statex $\mathcal{L}(\mathbf{Y}_{B},I(\mathbf{Y}_{B}),\hat{\mathbf{Y}}_{B},I(\hat{\mathbf{Y}}_{B}),\lambda)$    
         \State Update the parameters $\Theta_{GCN},\Theta_{PE}$ of model $M$ using stochastic gradient descent
         %\Statex $\mathcal{L}(\Theta_{PE},\Theta_{GCN})= \frac{1}{n_{batch}}\sum_{i=1}^{n_{batch}}(M_{\Theta_{PE},\Theta_{GCN}}(\mathbf{C}_{B},\mathbf{X}_{B})-\mathbf{Y}_{B})^{2}$
		\EndFor
		\State \textbf{return} $M$
		%\EndProcedure
	\end{algorithmic}
\end{algorithm}

We begin training by initializing our model $M$, for example a \textbf{PE-GCN}, with random weights and potential hyper-parameters (e.g., PE embedding dimension) and defining our optimizer. We then start the training cycle: At each training step, we first sample a minibatch $B$ of points from our training data. These points come as features $\mathbf{X}_{B}$, point coordinates $\mathbf{C}_{B}$ and outcome variables $\mathbf{Y}_{B}$. We construct a graph from spatial coordinates $\mathbf{C}_{B}$ using $k$-nearest-neighborhood, obtaining an adjacency matrix $\mathbf{A}_{B}$. Next we use $\mathbf{A}_{B}$ as spatial weight matrix to compute local Moran's I values $I(\mathbf{Y}_B)$ from $\mathbf{Y}_B$. As minibatches are randomly sampled, this creates a ``shuffled'' version of the metric. We then run inputs $\mathbf{X}_{B},\mathbf{C}_{B},\mathbf{A}_{B}$ through the two-headed model $M_{\Theta_{PE},\Theta_{GCN}}$ obtaining predictions $\hat{\mathbf{Y}}_{B},I(\hat{\mathbf{Y}}_{B})$. We then compute the loss $\mathcal{L}(\mathbf{Y}_{B},I(\mathbf{Y}_{B}),\hat{\mathbf{Y}}_{B},I(\hat{\mathbf{Y}}_{B}),\lambda)$, weighing the Moran's I auxiliary task according to weight parameter $\lambda$. Lastly, we use the loss $\mathcal{L}$ to update our model parameters $\Theta_{GCN},\Theta_{PE}$ according to stochastic gradient descent. Training is conducted for $tsteps$ after which the final model $M$ is returned.

\textbf{PE-GNN}, with any GNN backbone, helps to tackle many of the particular challenges of geographic data: While our approach still includes the somewhat arbitrary choice of $k$-nearest neighbors to define the spatial graph, the proposed positional encoder network is not bound by this restriction, as it does not operate on the graph. This enables a separate learning of context-aware embeddings for each coordinate, accounting for neighbors at any potential distance within the batch. While the spatial graph used still relies on pre-defined distance measure, the positional encoder embeds latitude and longitude values in a high-dimensional latent space. These high-dimensional coordinates are able to reflect spatial complexities much more flexibly and, added as node features, can communicate these throughout the learning process. Batched \textbf{PE-GNN} training is not conducted on a single graph, but a new graph consisting of randomly sampled training points at each iteration. As such, at different iterations, focus is put on the relationships between different clusters of points. This helps our method to generalize better, rather than just memorizing neighborhood structures. Lastly, the differing training batches also help us to compute a ``shuffled" version of the Moran's I metric, capturing autocorrelation at the same location for different (closer or more distant), random neighborhoods.

\begin{figure*}[!ht]
\begin{subfigure}{1\textwidth}
  \centering
  % include first image
  \includegraphics[scale=0.53]{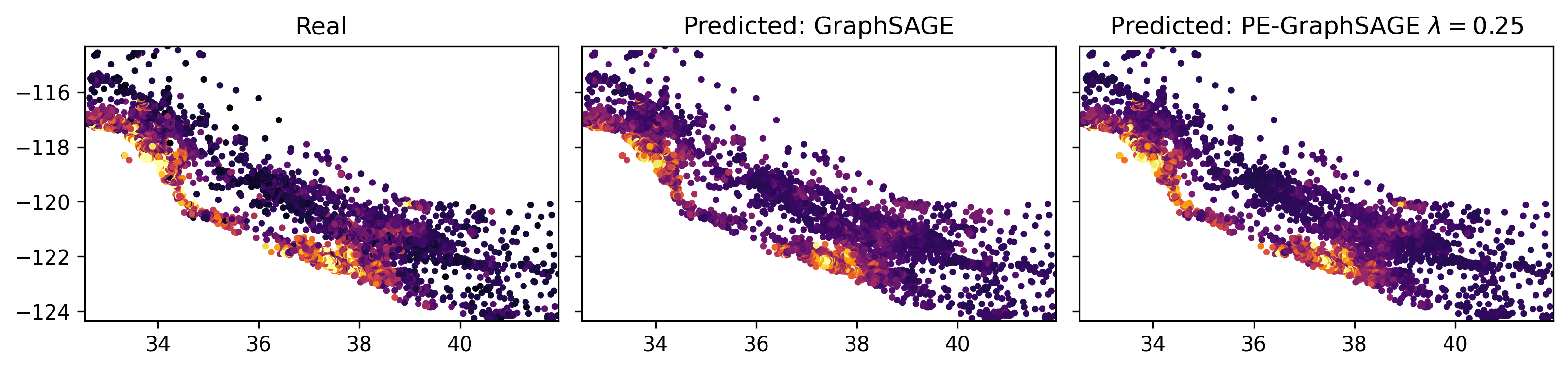}  
  \caption{Real values and predictions using GraphSAGE and PE-GraphSAGE.}
  \label{fig2:sub-first}
\end{subfigure}
\begin{subfigure}{1\textwidth}
  \centering
  % include second image
%  \includegraphics[width=1\linewidth]{cali_housing_mse_error.pdf}  
  \includegraphics[scale=0.5]{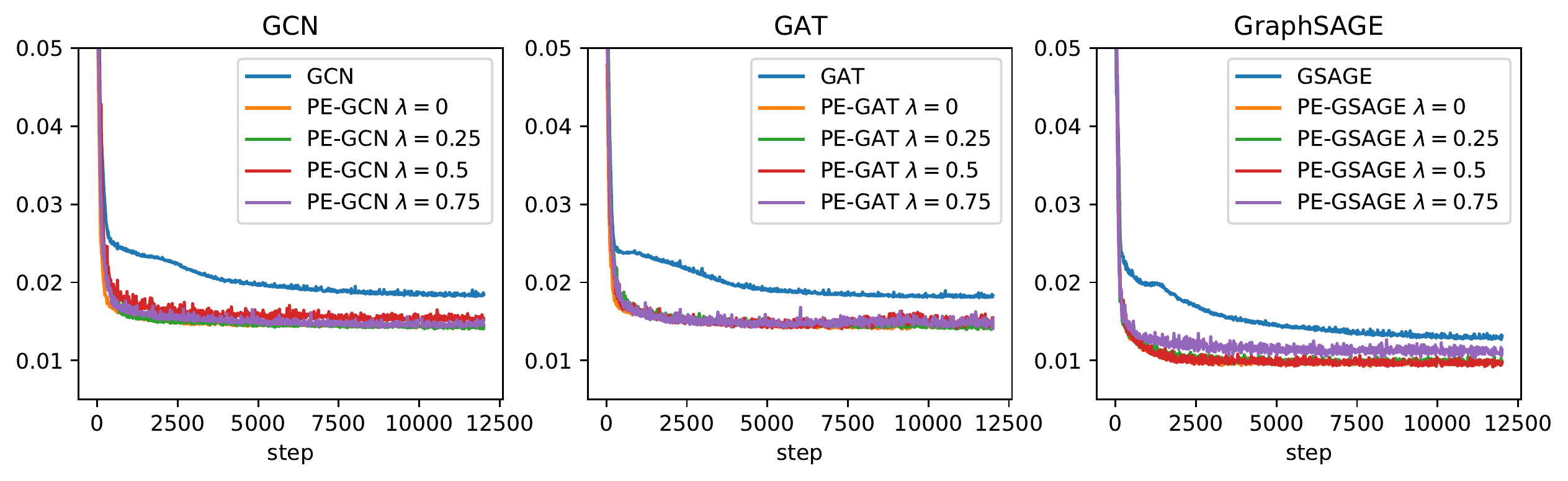}  
  \caption{Test error curves of GCN, GAT and GraphSAGE based models, measured by the MSE metric.}
  \label{fig2:sub-second}
\end{subfigure}
\caption{Visualizing predictive performance on the California Housing dataset.}
\label{fig:2}
\end{figure*}

%\begin{figure}[!ht]%{0.4\textwidth}
%\vskip -0.1in
%\centering
%\includegraphics[scale=0.41]{cali_housing_mse_error.pdf}
%\includegraphics{fig1.pdf}
%\vskip -0.1in
%\caption{Test}
%\label{fig:3}
%\vskip -0.15in
%\vskip -0.5in
%\end{figure}

\section{Experiments}

\begin{table*}[!ht]
\centering
\resizebox{12.5cm}{!}{%
\begin{tabular}{l|l|l|l|l|l|l|l|l}
\textbf{Model} & \multicolumn{2}{l}{\textbf{Cali. Housing}} & \multicolumn{2}{l}{\textbf{Election}} & \multicolumn{2}{l}{\textbf{Air Temp.}} & \multicolumn{2}{l}{\textbf{3d Road}} \\		
& MSE & MAE & MSE & MAE & MSE & MAE & MSE & MAE \\		
\hline
\hline
GCN \cite{Kipf2017} &  0.0558	&	0.1874  & 0.0034	&	0.0249 & 0.0225	&	0.1175 & 0.0169 & 0.1029 \\
PE-GCN $\lambda=0$ &  0.0161	&	\textbf{0.0868} & 0.0032	&	0.0241 & 0.0040	&	0.0432 & \textbf{0.0031}	&	\textbf{0.0396} \\
PE-GCN $\lambda=0.25$ & \textbf{0.0155}	&	0.0882 & 0.0032	&	\textbf{0.0236} & 0.0037	&	0.0417 & 0.0032	&	0.0416 \\
PE-GCN $\lambda=0.5$ &  0.0156	&	0.0885 & \textbf{0.0031}	&	0.0241 & \textbf{0.0036}	&	\textbf{0.0401} &  0.0033	&	0.0421 \\
PE-GCN $\lambda=0.75$ &  0.0160	&	0.0907 & \textbf{0.0031}	&	0.0240 & 0.0040	&	0.0429 &  0.0033	&	0.0424 \\
\hline
GAT \cite{Velickovic2018} &  0.0558	&	0.1877 & 0.0034	&	0.0249  & 0.0226	&	0.1165 & 0.0178	&	0.0998 \\
PE-GAT $\lambda=0$ &  \textbf{0.0159}	&	0.0918 & \textbf{0.0032}	&	\textbf{0.0234} & \textbf{0.0039}	&	0.0429 &  0.0060	&	0.0537 \\
PE-GAT $\lambda=0.25$ &  0.0161	&	\textbf{0.0867} & \textbf{0.0032}	&	0.0235  & 0.0040	&	\textbf{0.0417} & \textbf{0.0058}	&	\textbf{0.0530} \\
PE-GAT $\lambda=0.5$ & 0.0162	&	0.0897 & \textbf{0.0032}	&	0.0238  & 0.0045	&	0.0465 & 0.0061	&	0.0548 \\
PE-GAT $\lambda=0.75$ &  0.0162	&	0.0873 & \textbf{0.0032}	&	0.0237  & 0.0041	&	0.0429 & 0.0062	&	0.0562 \\
\hline
GraphSAGE \cite{Hamilton2017a} &  0.0558	&	0.1874 & 0.0034	&	0.0249  & 0.0274	&	0.1326 & 0.0180	&	0.0998 \\
PE-GraphSAGE $\lambda=0$ &  0.0157	&	0.0896 & \textbf{0.0032}	&	\textbf{0.0237} & 0.0039	&	0.0428 & 0.0060	&	\textbf{0.0534} \\
PE-GraphSAGE $\lambda=0.25$ &  \textbf{0.0097}	&	0.0664 & \textbf{0.0032}	&	0.0242  & 0.0040	&	0.0418 & 0.0059	&	\textbf{0.0534} \\
PE-GraphSAGE $\lambda=0.5$ &  0.0100	&	0.0682 & 0.0033	&	0.0239  & 0.0043	&	0.0461 & 0.0060	&	0.0536 \\
PE-GraphSAGE $\lambda=0.75$ &  0.0100	&	\textbf{0.0661} & \textbf{0.0032}	&	0.0241  & \textbf{0.0036}	&	\textbf{0.0399} & \textbf{0.0058}	&	0.0541 \\
\hline
KCN \cite{Appleby2020} &  	0.0292	&	0.1405	 & 	0.0367	&	0.1875	 &  0.0143	&	0.0927	& 0.0081	&	0.0758	 \\
PE-KCN $\lambda=0$ &   	0.0288	&	0.1274	 & 	0.0598	&	0.2387	 &  0.0648	&	0.2385	& \textbf{0.0025}	&	\textbf{0.0310}	 \\
PE-KCN $\lambda=0.25$ &  	0.0324	&	0.1380	 & 	0.0172	&	0.1246	  &  \textbf{0.0059}	&	\textbf{0.0593}	&	0.0037	&	0.0474 \\
PE-KCN $\lambda=0.5$ &  \textbf{0.0237}	&	\textbf{0.1117}   & 	0.0072	&	0.0714	  &  0.0077	&	0.0664	&	0.0077	&	0.0642 \\
PE-KCN $\lambda=0.75$ &  	0.0260	&	0.1194	 &  	\textbf{0.0063}	&	\textbf{0.0681}   &  0.0122	&	0.0852	&	0.0110	&	0.0755 \\
\hline
Approximate GP & 0.0353	&	0.1382 & 0.0031	& 	0.0348 & 0.0481	&	0.0498 &0.0080	& 0.0657 \\
Exact GP & 0.0132	&	0.0736 & 0.0022	&	0.0253 & 0.0084	&	0.0458 & - & -

\end{tabular}}
\caption{\textit{Spatial Interpolation:} Test MSE and MAE scores from four different datasets, using four different GNN backbones with and without our proposed architecture.} \label{tab:table1}
%\end{wraptable}
%\vskip -0.25in
\end{table*}

\begin{table*}[!ht]
\centering
\resizebox{9cm}{!}{%
\begin{tabular}{l|l|l|l|l|l|l}
\textbf{Model} & \multicolumn{2}{l}{\textbf{Cali. Housing}} & \multicolumn{2}{l}{\textbf{Election}} & \multicolumn{2}{l}{\textbf{Air Temp.}}  \\		
& MSE & MAE & MSE & MAE & MSE & MAE \\		
\hline
\hline
GCN &  0.0185 & 0.1006 & \textbf{0.0025} & \textbf{0.0211}  & 0.0225	& 0.1175\\
PE-GCN $\lambda=0$ &  \textbf{0.0143}	&	\textbf{0.0814}  &  0.0026	&	0.0213  &0.0040	&	0.0432\\
PE-GCN $\lambda=0.25$ &  \textbf{0.0143}	&	0.0816  &  0.0026	&	0.0213   & 0.0037	&	0.0417\\
PE-GCN $\lambda=0.5$ &  \textbf{0.0143}	&	0.0828  &  0.0027	&	0.0217   & \textbf{0.0036}	&	\textbf{0.0401}\\
PE-GCN $\lambda=0.75$ &  0.0147	&	0.0815  &  0.0027	&	0.0219   & 0.0040	&	0.0429\\
\hline
GAT &  0.0183	&	0.0969  &  \textbf{0.0024}	&	\textbf{0.0211} & 0.0226	&	0.1165 \\
PE-GAT $\lambda=0$ &  0.0144	&	0.0836  &  0.0028	&	0.0218 & \textbf{0.0039}	&	0.0429 \\
PE-GAT $\lambda=0.25$ &  \textbf{0.0141}	&	\textbf{0.0817}  &  0.0028	&	0.0219 & 0.0040	&	\textbf{0.0417}\\
PE-GAT $\lambda=0.5$ &  0.0155	&	0.0851  &  0.0030	&	0.0225 & 0.0045	&	0.0465 \\
PE-GAT $\lambda=0.75$ &  0.0145	&	0.0824  &  0.0029	&	0.0223 & 0.0041	&	0.0429 \\
\hline
G.SAGE &  0.0131	&	0.0798  &  \textbf{0.0007}	&	\textbf{0.0127} & 0.0219	&	0.1153 \\
PE-G.SAGE $\lambda=0$ &  0.0099	&	0.0667  &  0.0011	&	0.0154  & 0.0037	&	0.0422\\
PE-G.SAGE $\lambda=0.25$ &  \textbf{0.0098}	&	\textbf{0.0648}  &  0.0010	&	0.0152 &  \textbf{0.0029}	&	\textbf{0.0381}\\
PE-G.SAGE $\lambda=0.5$ &  \textbf{0.0098}	&	0.0679  &  0.0012	&	0.0157 &  0.0037	&	0.0445\\
PE-G.SAGE $\lambda=0.75$ &  0.0114	&	0.0766  &  0.0012	&	0.0152 & 0.0038	&	0.0459 \\
\hline
KCN &  	0.0292	&	0.1405	  &  	0.0367	&	0.1875	 &  0.0143	&	0.0927\\
PE-KCN $\lambda=0$ &  	0.0288	&	0.1274	  &  	0.0598	&	0.2387	 & 0.0648	&	0.2385 \\
PE-KCN $\lambda=0.25$ &  	0.0324	&	0.1380	  &  	0.0172	&	0.1246	&  \textbf{0.0059}	&	\textbf{0.0593}\\
PE-KCN $\lambda=0.5$ & 	\textbf{0.0237}	&	\textbf{0.1117}	 &  	0.0072	&	0.0714	 & 0.0077	&	0.0664 \\
PE-KCN $\lambda=0.75$ &  	0.0260	&	0.1194	  &  	\textbf{0.0063}	&	\textbf{0.0681}	 &  0.0122	&	0.0852\\
\hline
Approximate GP & 0.0195	&	0.1008 & 0.0050	&	0.0371 & 0.0481	&	0.0498 \\
Exact GP & 0.0036	&	0.0375 & 0.0006	&	0.0139 & 0.0084	&	0.0458

\end{tabular}}
\caption{\textit{Spatial Regression:} Test MSE and MAE scores from three different datasets, using four different GNN backbones with and without our proposed architecture.} \label{tab:table2}
%\end{wraptable}
%\vskip -0.25in
\end{table*}

\begin{figure*}[!ht]
    \centering
%\begin{wrapfigure}{L}{0.7\textwidth}
    \begin{subfigure}{0.9\textwidth}
      \centering
      % include first image
      \includegraphics[scale=0.45]{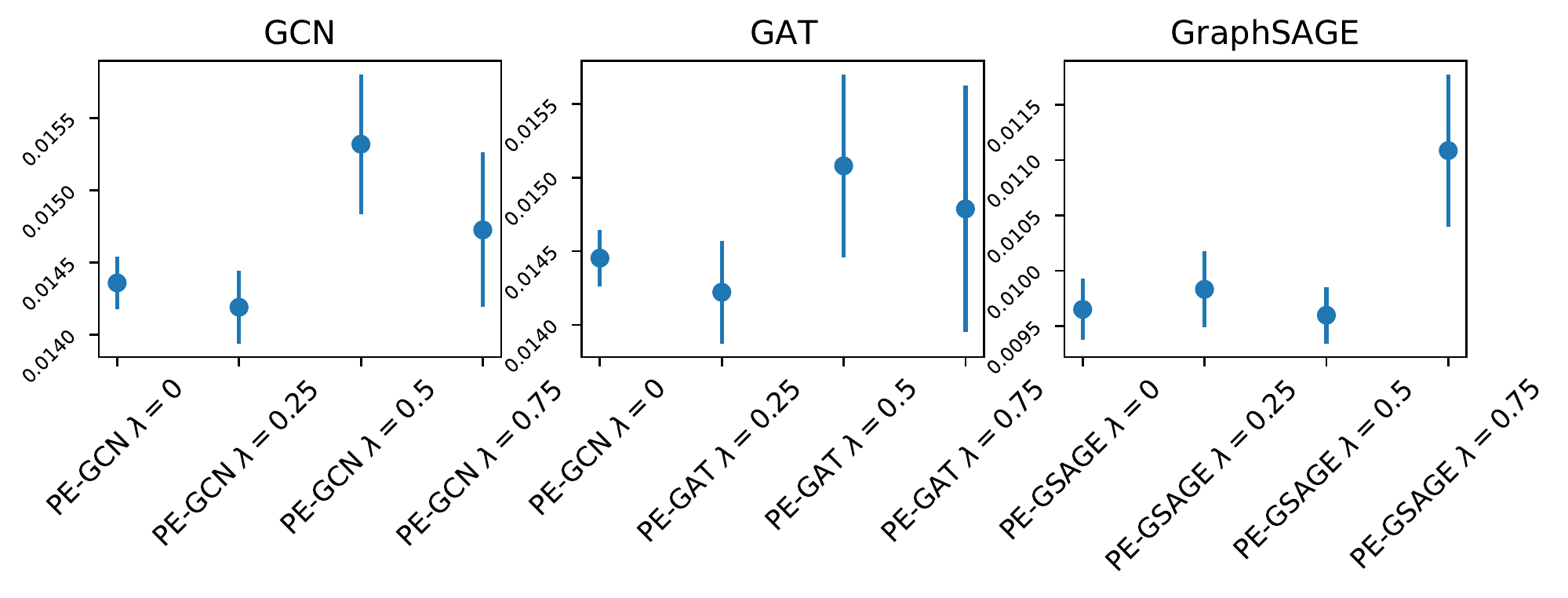}  
      \caption{California Housing.}
      \label{fig3:sub-first}
    \end{subfigure}
    \begin{subfigure}{0.9\textwidth}
      \centering
      % include second image
    %  \includegraphics[width=1\linewidth]{cali_housing_mse_error.pdf}  
      \includegraphics[scale=0.45]{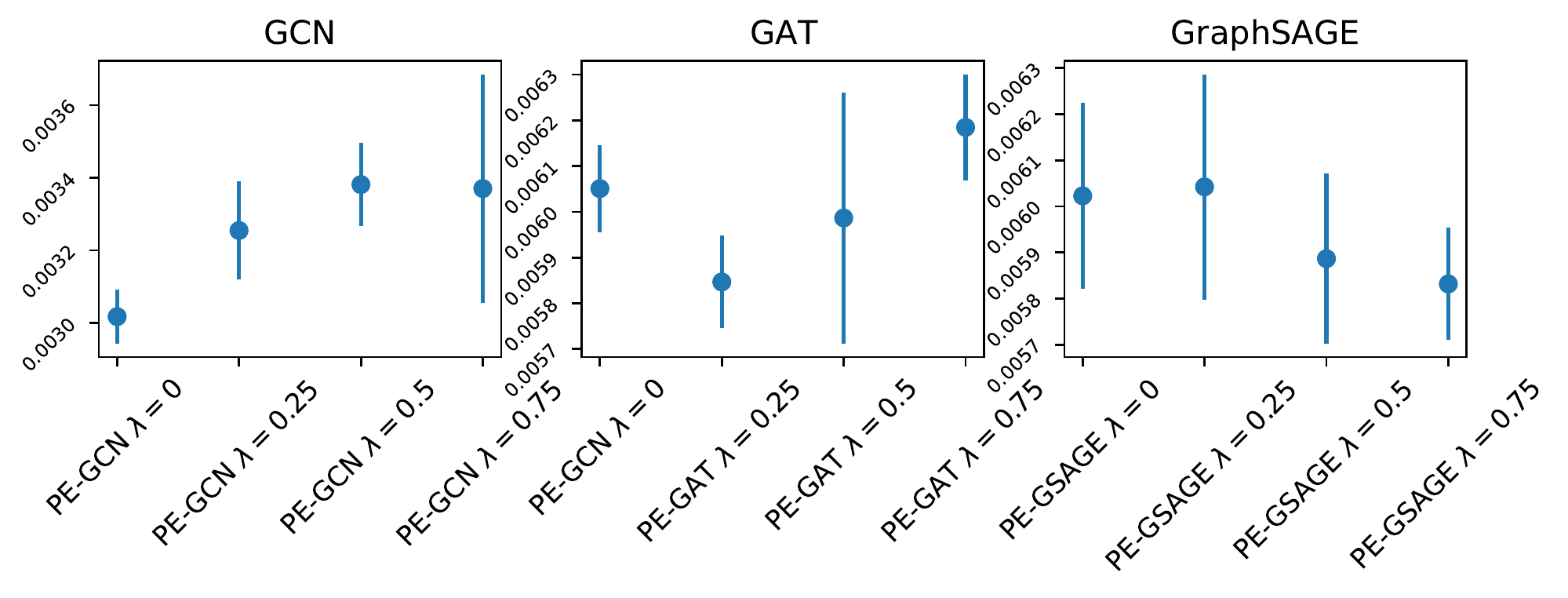}  
      \caption{3d Road.}
      \label{fi3:sub-second}
    \end{subfigure}
    \caption{MSE bar plots of mean performance and $2\sigma$ confidence intervals obtained from 10 different training checkpoints.}
    \label{fig:3}
%\end{wrapfigure}
\end{figure*}

\subsection{Data}

We evaluate \textbf{PE-GNN} and baseline competitors on four real-world geographic datasets of different spatial resolutions (regional, continental and global):\\
\textit{California Housing:} This dataset contains the prices of over $20,000$ California houses from the 1990 U.S.~census \citep{KelleyPace2003}. The regression task at hand is to predict house prices $y$ using features $\mathbf{x}$ (e.g., house age, number of bedrooms) and location $\mathbf{c}$. California housing is a standard dataset for assessment of spatial autocorrelation.\\
\textit{Election:}This dataset contains the election results of over $3,000$ counties in the United States \citep{Jia2020}. The regression task here is to predict election outcomes $y$ using socio-demographic and economic features (e.g., median income, education) $\mathbf{x}$ and county locations $\mathbf{c}$.  \\
\textit{Air temperature:}The air temperature dataset \citep{Hooker2018} contains the coordinates of ~$3,000$ weather stations around the globe. For this regression task we seek to predict mean temperatures $y$ from a single node feature $x$, mean precipitation, and location $c$. \\
\textit{3d Road:}The 3d road dataset \citep{Kaul2013} provides $3$-dimensional spatial co-ordinates (latitude, longitude, and altitude) of the road network in Jutland, Denmark. The dataset comprises over $430,000$ points and can be used for interpolating altitude $y$ using only latitude and longitude coordinates $\mathbf{c}$ (no node features $x$). % Inferring the third spatial dimension from two observed spatial dimensions can also be thought of as point-cloud completion or shape-from-x problem.

\subsection{Experimental setup}

We compare \textbf{PE-GNN} with four different graph neural network backbones: The original GCN formulation \citep{Kipf2017}, graph attention mechanisms (GAT) \citep{Velickovic2018} and GraphSAGE \citep{Hamilton2017a}. We also use Kriging Convolutional Networks (KCN) \citep{Appleby2020}, which differs from GCN primarily in two ways: it transforms the distance-weighted adjacency matrix $\mathbf{A}$ using a Gaussian kernel and adds the outcome variable and features of neighboring points to the features of each node. Test set points can only access neighbors from the training set to extract these features. We compare the naive version of all these approaches to the same four backbone architectures augmented with our \textbf{PE-GNN} modules. Beyond GNN-based approaches, we also compare \textbf{PE-GNN} to the most popular method for modeling continuous spatial data: Gaussian processes. For all approaches, we compare a range of different training settings and hyperparameters, as discussed below.

To allow for a fair comparison between the different approaches, we equip all models with the same architecture, consisting of two GCN / GAT / GraphSAGE layers with ReLU activation and dropout, followed by linear layer regression heads. The KCN model also uses GCN layers, following the author specifications. We found that adding additional layers to the GNNs did not increase their capacity for processing raw latitude / longitude coordinates. We test four different auxiliary task weights $\lambda= \{0,0.25.0.5,0.75\}$, where $\lambda=0$ implies no auxiliary task. Spatial graphs are constructed assuming $k=5$ nearest neighbors, following rigorous testing. This also confirms findings from previous work \citep{Appleby2020,Jia2020}. We include a sensitivity analysis of the $k$ parameter and different batch sizes in our results section. Training for the GNN models is conducted using PyTorch \citep{Paszke2019} and PyTorch Geometric \citep{Fey2019}. We use the Adam algorithm to optimize our models \citep{Kingma2015} and the mean squared error (MSE) loss. Gaussian process models (exact and approximate) are trained using GPyTorch \citep{Gardner2018}. Due to the size of the dataset, we only provide an approximate GP result for 3d Road. All training is conducted on single CPU. On the Cali.~Housing dataset ($n>20,000$) training times for one step (no batched training) are as follows: PE-GCN = $0.23s$ (with aux. task $0.24s$), PE-GAT = $0.38s$, PE-GraphSAGE = $0.33s$, PE-KCN = $0.41$, exact GP = $0.77s$. Results are averaged over $100$ training steps. The code for \textbf{PE-GNN} and our experiments can be accessed here: \url{https://github.com/konstantinklemmer/pe-gnn}.

\subsection{Results}

\subsubsection{Predictive performance}

We test our methods on two tasks: \textit{Spatial Interpolation}, predicting outcomes from spatial coordinates alone, and \textit{Spatial Regression}, where other node features are available in addition to the latitude / longitude coordinates.
The results of our experiments are shown in Table \ref{tab:table1} and \ref{tab:table2}. For all models, we provide mean squared error (MSE) and mean absolute error (MAE) metrics on held-out test data. For the \textit{spatial interpolation} task, we observe that the \textbf{PE-GNN} approaches consistently and vastly improve performance for all four backbone architectures across the California Housing, Air Temperature and 3d Road datasets and, by a small margin, for the Election dataset. For the \textit{spatial regression} task, we observe that the \textbf{PE-GNN} approaches consistently and substantially improve performance for all four backbone architectures on the California Housing and Air Temperature datasets. Performance remains unchanged or decreases by very small margins in the Election dataset, except for the KCN backbone which benefits tremendously from the \textbf{PE-GNN} approach, particularly with auxiliary tasks.

Generally, \textbf{PE-GNN} substantially improves over baselines in regression and interpolation settings. Most of the improvement can be attributed to the positional encoder, however the auxiliary task learning also has substantial beneficial effects in some settings, especially for the KCN models. The best setting for the task weight hyperparameter $\lambda$ seems to heavily depend on the data, which confirms findings by \cite{Klemmer2021a}.
To our knowledge, \textbf{PE-GNN} is the first GNN-based learning approach that can compete with Gaussian Processes on simple spatial interpolation baselines, though especially exact GPs still sometimes have the edge. \textbf{PE-GNN} is substantially more scalable than exact GPs, which rely on expensive pair-wise distance calculations across the full training dataset. Due to this problem, we do not run an exact GP baseline for the high-dimensional 3d Road dataset. For KCN models, we observe a proneness to overfitting. As the authors of KCN mention, this effect diminishes in large enough data domains \citep{Appleby2020}. For example, KCNs are the best performing method on the 3d Road dataset--by far our largest experimental dataset. Here, we also observe that in cases when KCN learns well, \textbf{PE-KCN} can still improve its performance. The KCN experiments also highlight the strongest effects of the Moran's I auxiliary tasks: In cases when KCN overfits (Election, Cali. Housing datasets), \textbf{PE-KCN} without auxiliary task ($\lambda=0$) is not sufficient to overcome the problem. However, adding the auxiliary task can mitigate most of the overfitting issue. This directly confirms a theory of \cite{Klemmer2021a} on the beneficial effects of auxiliary learning of spatial autocorrelation. Regarding the question of spatial scale, we find no systemic variation in \textbf{PE-GNN} performance between applications with regional (California Housing, 3d Road), continental (Election) and global (Air Temperature) spatial coverage. \textbf{PE-GNN} performance depends on the difficulty of the task at hand and the complexity of present spatial dependencies.

%Throughout our experiments, \textbf{PE-GNN} provides a powerful and flexible framework for learning with geogpraphic coordinates. The positional encoder allows for the learning of spatial context features to inform the downstream task, while the Moran's I auxiliary task helps to overcome potential overfitting issues. 

We also assess the robustness of \textbf{PE-GNN} training cycles. Figure \ref{fig:3} highlights the confidence intervals of \textbf{PE-GNN} models with GCN, GAT and GraphSAGE backbones trained on the California Housing and 3d Road datasets, obtained from 10 different training cycles. We can see that training runs exhibit only little variability. These findings thus confirm that \textbf{PE-GNN} can consistently outperform naive GNN baselines.

\begin{figure}[ht!]%{0.4\textwidth}
%\vskip -0.1in
\centering
\includegraphics[scale=0.5]{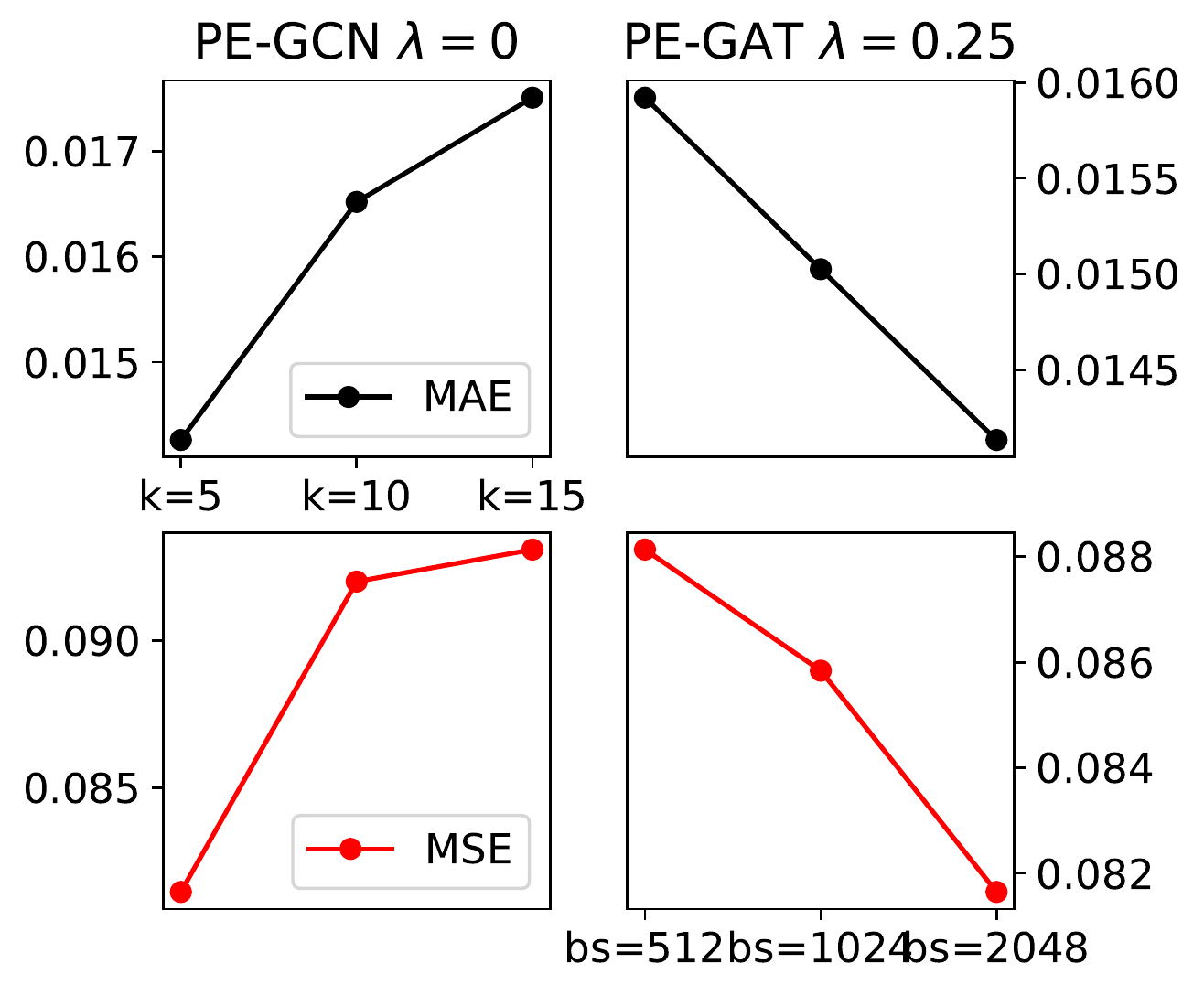}
%\includegraphics{fig1.pdf}
%\vskip -0.1in
\caption{Predictive performance of PE-GCN and PE-GAT models on the California Housing dataset, using different values of $k$ for constructing nearest-neighbor graphs and different batch sizes (bs).}
\label{fig:4}
%\vskip -0.15in
%\vskip -0.5in
\end{figure}

\subsubsection{Sensitivity analyses}

Figure \ref{fig:4} highlights some results from our sensitivity analyses with the $k$ and $n_{batch}$ (batch size) parameters. After rigorous testing, we opt for $k=5$-NN approach to create the spatial graph and compute the shuffled Moran's I across all models. We chose $n_{batch}=2048$ for Cali. Housing and 3d Road datasets and $n_{batch}=1024$ for the Election and Air Temperature datasets. Note that while our experiments focus on batched training to highlight the applicability of \textbf{PE-GNN} to high-dimensional geospatial datasets, we also tested our approach with non-batched training on the smaller datasets (Election, Air Temperature, California Housing). We found only marginal performance differences between these settings.

\begin{figure}[ht!]%{0.4\textwidth}
%\vskip -0.1in
\centering
\includegraphics[scale=0.6]{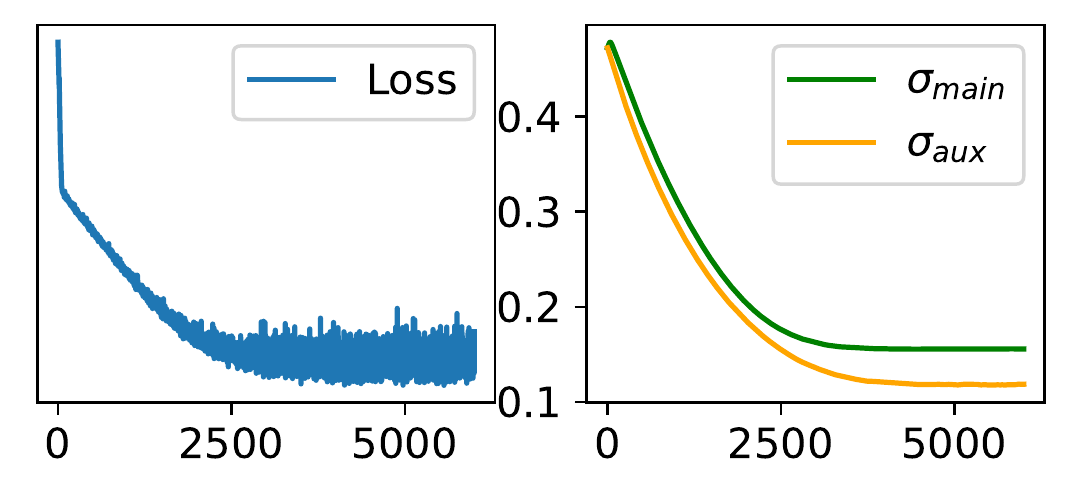}
%\includegraphics{fig1.pdf}
%\vskip -0.1in
\caption{Automatic learning of loss weights via task uncertainty on the Air Temp. dataset with PE-GCN. The left graphic shows the training loss (MSE), while the right graphic shows the main and auxiliary task weight parameters $\sigma_{main}$ and $\sigma_{aux}$. The training steps are given on the x-axis.}
\label{fig:5}
%\vskip -0.15in
%\vskip -0.5in
\end{figure}

\subsubsection{Learning auxiliary loss weights using task uncertainty}

Lastly, following work by \citet{Cipolla2018} and \citet{Klemmer2021a}, we provide an intuition for automatically selecting the Moran's I auxiliary task weights using task uncertainty. This eliminates the need to manually tune and select the $\lambda$ parameter. The approach first proposed by \citet{Cipolla2018} formalizes the idea by first defining a probabilistic multi-task regression problem with a main and auxiliary task as:

\begin{equation}
\begin{aligned}
    p(\hat{\mathbf{Y}}_{main}, \hat{\mathbf{Y}}_{aux} | f(\mathbf{X})) = p(\hat{\mathbf{Y}}_{main} | f(\mathbf{X})) p(\hat{\mathbf{Y}}_{aux} | f(\mathbf{X}))
\end{aligned}
\end{equation}

with $\hat{\mathbf{Y}}_{main}, \hat{\mathbf{Y}}_{aux}$ giving the main and auxiliary task predictions. Following maximum likelihood estimation, the regression objective function is given as $\min \mathcal{L}(\sigma_{main},\sigma_{aux})$:

\begin{equation}
\begin{aligned}
    & = -\log p(\hat{\mathbf{Y}}_{main},\hat{\mathbf{Y}}_{aux} | f(\mathbf{X})) \\
    & = \frac{1}{2 \sigma_{main}^{2}} \mathcal{L}_{main} + \frac{1}{2 \sigma_{aux}^{2}} \mathcal{L}_{aux} + \\
    & (\log \sigma_{main} + \log \sigma_{aux}), 
\end{aligned}
\end{equation}

with $\sigma_{main}$ and $\sigma_{aux}$ defining the model noise parameters. By minimizing this objective, we learn the relative weight or contribution of main and auxiliary task to the combined loss. The last term of the loss prevents it from moving towards infinity and acts as a regularizer. While this approach performs equally compared to a well selected $\lambda$ parameter, it eliminates the need to manually tune and select $\lambda$. Figure \ref{fig:5} highlights the learning of $\sigma_{main}$ and $\sigma_{aux}$ loss weights using \textbf{PE-GCN} and the Air Temperature dataset.

\section{Conclusion}
With \textbf{PE-GNN}, we introduce a flexible, modular GNN-based learning framework for geographic data. \textbf{PE-GNN} leverages recent findings in embedding spatial context into neural networks to improve predictive models. Our empirical findings confirm a strong performance. This study highlights how domain expertise can help improve machine learning models for applications with distinct characteristics. We hope to build on the foundations of \textbf{PE-GNN} to develop further methods for geospatial machine learning.

% \section*{Acknowledgements}
% The authors gratefully acknowledge funding from the UK Engineering  and  Physical  Sciences  Research  Council, the EPSRC Centre for Doctoral Training in Urban Science (EP-SRC grant no. EP/L016400/1).

\bibliographystyle{ACM-Reference-Format}
\bibliography{aaai22_konstantin, aaai22_nathan}

\appendix

%EMPTY

\end{document}